\documentclass[Journal,twocolumn,12pt]{IEEEtran}

\usepackage{epstopdf}

\usepackage{graphicx}
\usepackage{amssymb}
\usepackage{amsmath}
\usepackage{subfigure}

\usepackage{caption}

\newcommand{\comment}[1]{}

     \newcommand{\qed}{\nobreak \ifvmode \relax \else
           \ifdim\lastskip<1.5em \hskip-\lastskip
           \hskip1.5em plus0em minus0.5em \fi \nobreak
           \vrule height0.75em width0.5em depth0.25em\fi}
\begin{document}


\title{Multi-Camera Occlusion and Sudden-Appearance-Change Detection Using Hidden Markovian Chains}

\author{\authorblockN{Xudong Ma \\}
\authorblockA{Pattern Technology Lab LLC, U.S.A.\\
Email: xma@ieee.org} }

\maketitle

\begin{abstract}

This paper was originally submitted to \mbox{Xinova} as a response to a Request for Invention (RFI) on new event monitoring methods. In this paper, a new object tracking algorithm using multiple cameras for surveillance applications is proposed. The proposed system can detect sudden-appearance-changes and occlusions using a hidden Markovian statistical model. The experimental results confirm that our system detect the sudden-appearance changes and occlusions reliably.


\end{abstract}

\section{Introduction}

\label{section_introduction}

In surveillance applications, the ability of long-term tracking a certain person or object from security cameras is highly desirable. Usually the security personnel can label a suspicious person in the video. The tracking system can then track this suspicious person in the videos without further human inputs for minutes or hours.

One situation that the visual tracking systems for the surveillance applications may deal with frequently is the sudden-appearance-changes of the being tracked person or object. For example, the suspicious person may take off his/her jacket, pull up his/her hood, or abandon some luggage, in order to fool the surveillance system.  Such sudden-appearance-changes are suspicious activities and  usually should be reported to the security personnel in real-time.

Unfortunately, it is usually difficult for computer vision algorithms to distinguish between a sudden-appearance-change to occlusions. For many surveillance applications, the surveillance scenes are usually crowded with many people. Thus, occlusions may happen very frequently. Without the ability of distinguishing between sudden-appearance-changes to occlusions, the visual tracking systems may generate a large number of false alarms.

Detecting occlusions correctly is also important for enhancing the reliability of visual tracking algorithms. It is well-known that for visual tracking, there exists a so called ``high-adaptability-to-drifting-resistance trade-off'' problem. The problem relates to how much we should update our models for the being-tracked person or object on-the-fly. If we do not update the models much, then in the case that the appearance of the being tracked person changes rapidly, there is a high risk of tracking loss. If we always update the models very rapidly, then in the case of occlusions, the model may be tuned to occulders. And these wrongly tuned models may result in the so called drifting. That is, the tracking algorithm may start to track the occulders instead. In Fig. \ref{exp1} of section \ref{section_simulation_result}, we actually show an example of such drifting phenomena.

In this paper, we propose a new occlusion and appearance-change detection method. The proposed real-time visual tracking system uses multiple surveillance cameras. Initially, the security personnel provides one bounding box of the suspicious person for each video frame sequence. The visual tracking system then tracks the where-about of the suspicious person in real-time. If a sudden-appearance-change of the suspicious person is detected, then the visual tracking system would raise an alarm signal immediately.

Our method uses both generative and discriminative models for the video frame streams. For each camera, one discriminative model is maintained for discriminating the image patches that contain the being-tracked person to the image patches that do not contain the being-tracked person. Similarly, one generative model is maintained for each camera. In this paper, we use a recently proposed compressive sensing and naive Bayes based classier in \cite{zzy12} as the discriminative model. We use linear sub-space models as the generative models. That is, we assume that the image patches containing the being-tracked person from several adjacent video frames all are vectors within a certain affine sub-space.

A center component of our method is a hidden Markovian model for the prediction errors of the generative models. That is, whenever a new video frame is received, the new image patch containing the being-tracked person is predicted from the previous such image patches. The hidden Markovian model thus contains a visible part and a hidden part. The visible part contains the observed prediction errors. And the hidden part contains random variables $O_1(t),  O_2(t), \ldots, O_N(t)$ and $S(t)$. The binary random variable $O_n(t)$ denotes whether an occlusion has occurred for the $n$-th camera at time $t$. And the binary random variable $S(t)$ denotes whether a sudden-appearance-change of the being-tracked person has occurred at time $t$.

We assume some parametric probability distributions for the hidden Markovian model. The probabilities of $O_n(t)$ and $S(t)$ are estimated from the observed prediction error $z_n(t)$ by using sequential Bayesian estimation. An alarm signal may be raised, if we detect a high probability of $S(t)=1$, a sudden-appearance-change occurred. The estimated probabilities of $O_n(t)$ and $S(t)$ are also used for adjusting the learning rates of the discriminative and generative models. It should be intuitively clear that any appearance-change of the being tracked objects may result in significant prediction errors at the same time at all the cameras, but occlusions usually result in significant prediction errors only at a few cameras. Their probabilities can thus be estimated accordingly.

Please note that the above hidden Markovian statistical model is the centerpiece of the proposed occlusion and sudden-appearance-change detection methods. The statistical model works also well with other discriminative and generative models.

There is a large literature on visual tracking algorithms, such as \cite{CRM03}, \cite{A04}, \cite{A07}. We have no intention here to provide a throughout survey on the general visual tracking algorithms. The approaches of using multiple cameras for visual tracking have become attractive in the recent years, due to the availability of large quantities of low-cost commodity cameras. There are some previous discussions on visual tracking using multiple cameras. In \cite{QKRBS07} \cite{CA99}, approaches are discussed, where the responsibilities of tracking may be passed from one camera to another camera.  In \cite{BLB07}, from each camera,  a statistical estimation of the location of the being tracked person or object is obtained independently. The independent estimations are then fused into a joint location estimation. In \cite{EM08}, video frames from multiple cameras are projected on a reference frame using homography transforms, such that  the signals corresponding to the being tracked person or object may be added constructively.

The rest of the paper is organized as follows. In Section \ref{section_basic_scheme}, we discuss the proposed visual tracking system and the hidden Markovian model. In Section \ref{section_online_estimation}, we present the sequential Bayesian estimation methods for the hidden Markovian model. Simulation results of the proposed method is provided in Section \ref{section_simulation_result}. Finally, some concluding remarks are presented in Section \ref{sec_conclusion}.

\section{Visual Tracking System and Hidden Markovian Model}
\label{section_basic_scheme}

\begin{figure}[h]
 \centering
 \includegraphics[width=3in]{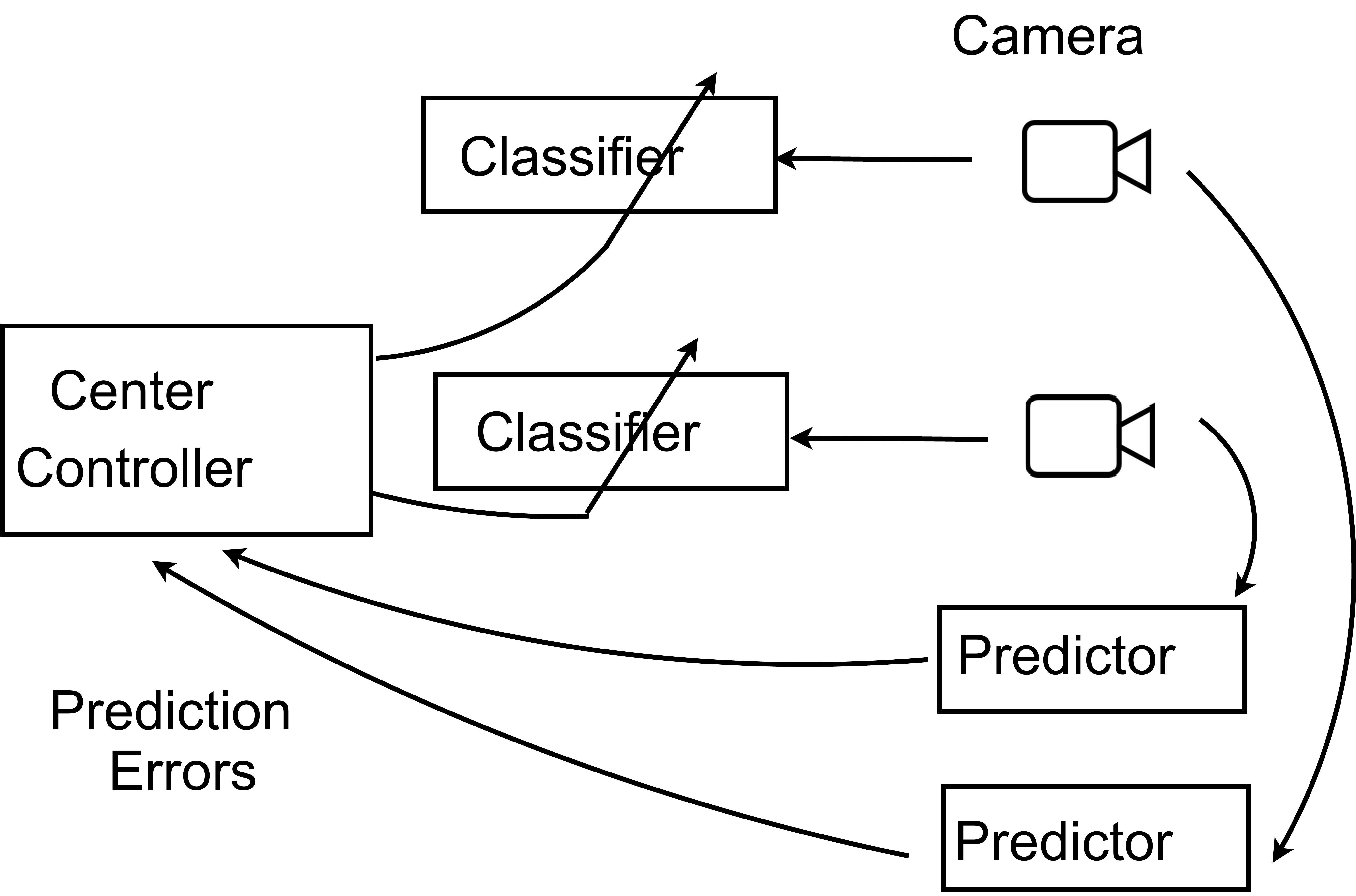}
 \caption{Block diagram of the visual tracking system}
 \label{basic_system_model}
\end{figure}

A block diagram of the proposed visual tracking system is shown in Fig. \ref{basic_system_model}. The system uses multiple cameras (only 2 are shown here). The system starts tracking a suspicious person, after the security personnel provides one bounding box of the suspicious person for each camera. For each camera, there is a real-time tracking sub-system as in \cite{zzy12} (shown as classifiers in Fig. \ref{basic_system_model} ). All such real-time tracking sub-systems work almost independently, except that their learning parameters $\lambda$ are controlled by the center controller.

The learning parameters $\lambda$ control how much each individual real-time compressive tracking sub-system should update the discriminative model after receiving each new video frame. As the being-tracked person changes his/her pose, orientation etc., the appearance of the person may change smoothly. Thus, each compressive tracking sub-system may update its discriminative model according to each newly observed video frame. The parameter $\lambda$ is a real number between $0$ and $1$. If $\lambda=1$, then the compressive tracking sub-system does not update its model. If $0<\lambda<1$, then the compressive tracking sub-system updates the discriminative model using a combination of past observed video frames and the newly observed video frame.

Our proposed method adjusts the learning parameters $\lambda$ based on the probabilities of sudden-appearance-change ${\mathbb P}[S(t)]$ and the probabilities of occlusions ${\mathbb P}\left[O_n(t)\right]$. Let us assume that we use $N$ security cameras. Suppose at each time $t = 1,2, \ldots$, we receive one video frame $x_n(t)$ at each camera, where $1\leq n \leq N$. Each tracking sub-system then finds one image patch $Y_n(t)$ that contains the being-tracked person, where $Y_n(t)$ is a column vector. That is, $Y_n(t)$ is the vector obtained by stacking the pixels in the image patch.

For each camera, we use one generative model (shown as predictors in Fig. \ref{basic_system_model}). Each predictor maintains an estimation of an affine subspace ${Space}_n(t)$, such that all the past observed $Y_n(t-1)$, $Y_n(t-2)$,$\ldots$ roughly lie within this affine space. We can then define a prediction error $Z_n(t)$ as the distance of the newly observed $Y_n(t)$ to this affine space  ${Space}_n(t)$. Note that there exist efficient algorithms for computing the affine space  ${Space}_n(t)$, such as in  \cite{ll00}.

We may then estimate the probabilities of sudden-appearance-change ${\mathbb P}\left[S(t)\right]$ and the probabilities of occlusions ${\mathbb P}\left[O_n(t)\right]$ from $Z_n(t)$ based on the following hidden Markovian model. We assume that the probability density function of $Z_n(t)$
\begin{align}
& {\mathbb P}[Z_n(t)| S(t),O_n(t)] \notag \\
& = \left\{
\begin{array}{ l l }
  \exp(- x / \mu)/\mu, & \mbox{ if } S(t) = O_n(t) = 0 \\
  1/M, & \mbox{ otherwise } \notag
\end{array}
\right.
\end{align}
where, $S(t)=1$ indicates that a sudden-appearance-change has occurred at time $t$ and $O_n(t)=1$ indicates that an occlusion has occurred at the $n$-th camera at time $t$. In other words, if $S(t)=O_n(t)=0$, then the prediction error $Z_n(t)$ is exponentially distributed. Otherwise, the prediction error $Z_n(t)$ is uniformly distributed between $0$ and $M$. We further assume that $Z_n(t)$ and $Z_m(t)$ are statistically independent conditioned on $S(t)$ and $O_1(t), \ldots, O_N(t)$, if $n\neq m$. We assume that each random process $O_n(1), O_n(2),\ldots, O_n(t)$ is a Markovian random process. Similarly, We also assume that the random process $S(1), S(2),\ldots, S(t)$ is Markovian.

We may then use the Bayesian decision methods to detect occlusions at each camera $O_n(t)$ and the sudden appearance change of the being tracked person or object $S(t)$ by computing the probabilities
\begin{align}
& {\mathbb P}[S(t)=1 | Z_{1:N}(1:t) ] \notag \\
& = {\mathbb P}[S(t)=1 | Z_1(1:t), Z_2(1:t), \ldots, Z_N(1:t) ] \notag \\
& {\mathbb P}[O_n(t)=1 | Z_{1:N}(1:t) ] \notag \\
& = {\mathbb P}[O_n(t)=1 | Z_1(1:t), Z_2(1:t), \ldots, Z_N(1:t) ] \notag
\end{align}
where $Z_n(1:t)$ denotes the collection of observed prediction errors $Z_n(1), Z_n(2), \ldots, Z_n(t)$, and $Z_{1:N}(1:t)$ denote the collection of variables $Z_1(1:t), \ldots, Z_N(1:t)$.
We show in Section \ref{section_online_estimation} that these probabilities can be recursively calculated in a very efficient way.

The proposed method raises an alarm signal, whenever the probability ${\mathbb P}[S(t)=1 | Z_{1:N}(1:t) ] $ goes over a certain threshold. The method may also adjust the learning parameters $\lambda$ of the compressive tracking sub-systems according to the probability ${\mathbb P}[O_n(t)=1 | Z_{1:N}(1:t) ]$. For example, we may set $\lambda=1$, whenever the probability ${\mathbb P}[O_n(t)=1 | Z_{1:N}(1:t) ]$ is greater than $0.5$.

\section{Recursive Bayesian Estimation}

\label{section_online_estimation}

In this section, we derive a recursive formula for calculating ${\mathbb P}\left[S(t), O_{1:N}(t)|Z_{1:N}(1:t)\right]$ from ${\mathbb P}\left[S(t-1), O_{1:N}(t-1)|Z_{1:N}(1:t-1)\right]$ in Eq. \ref{recursive_equation}, where (a) follows from the Markovian properties of the statistical model.

\begin{figure*}
\begin{align}
& {\mathbb P}\left[S(t), O_{1:N}(t)|Z_{1:N}(1:t)\right]  = \frac{
{\mathbb P}\left[S(t), O_{1:N}(t), Z_{1:N}(1:t)\right]
}{
{\mathbb P}\left[Z_{1:N}(1:t)\right]} \propto
 {\mathbb P}\left[S(t), O_{1:N}(t), Z_{1:N}(1:t)\right] \notag \\
& \propto {\mathbb P}\left[S(t), O_{1:N}(t), Z_{1:N}(t) | Z_{1:N}(1:t-1) \right] \notag \\
& = \sum_{S(t-1), O_{1:N}(t-1)} {\mathbb P}\left[S(t), O_{1:N}(t), S(t-1), O_{1:N}(t-1),  Z_{1:N}(1:t) | Z_{1:N}(1:t-1)\right] \notag \\
& = \sum_{S(t-1), O_{1:N}(t-1)} {\mathbb P}\left[S(t-1), O_{1:N}(t-1), | Z_{1:N}(1:t-1)\right] \notag \\
& \hspace{0.5in} \times  {\mathbb P}\left[S(t), O_{1:N}(t) | S(t-1), O_{1:N}(t-1),  Z_{1:N}(1:t-1)\right] \notag \\
& \hspace{0.5in} \times {\mathbb P}\left[z_{1:N}(t) | S(t), O_{1:N}(t) , S(t-1), O_{1:N}(t-1),  Z_{1:N}(1:t-1)\right] \notag \\
& \stackrel{(a)}{=} \sum_{S(t-1), O_{1:N}(t-1)} {\mathbb P}\left[S(t-1), O_{1:N}(t-1), | Z_{1:N}(1:t-1)\right] \notag \\
& \hspace{0.5in} \times  {\mathbb P}\left[S(t), O_{1:N}(t) | S(t-1), O_{1:N}(t-1)\right]  {\mathbb P}\left[z_{1:N}(t) | S(t), O_{1:N}(t) \right] \label{recursive_equation}
\end{align}
\end{figure*}

\section{Experimental Result}
\label{section_simulation_result}

We use one of the PETS 2007 (Tenth IEEE International Workshop on
Performance Evaluation of Tracking and Surveillance) data-sets (available from  http://www.cvg.reading.ac.uk/PETS2007/data.html). The data-set consists of 12000 video frames for 4 cameras, 3000 video frames for each camera. A suspicious person enters the scene at roughly frame 500 and drops and leaves behind his backpack at roughly frame 850.

\begin{figure*}[h!]
\begin{tabular}{|c|c|c|}
      \hline
      \includegraphics[width=2in]{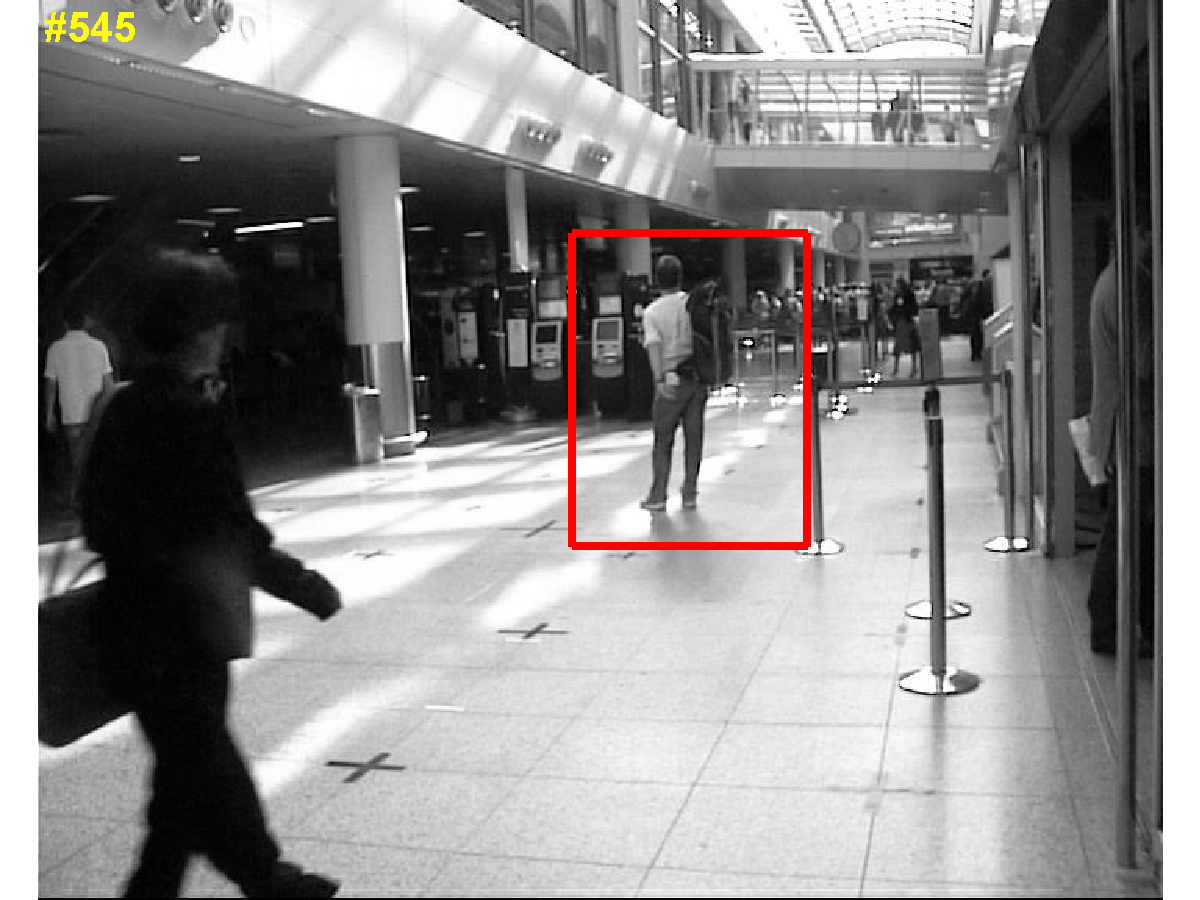} &
      \includegraphics[width=2in]{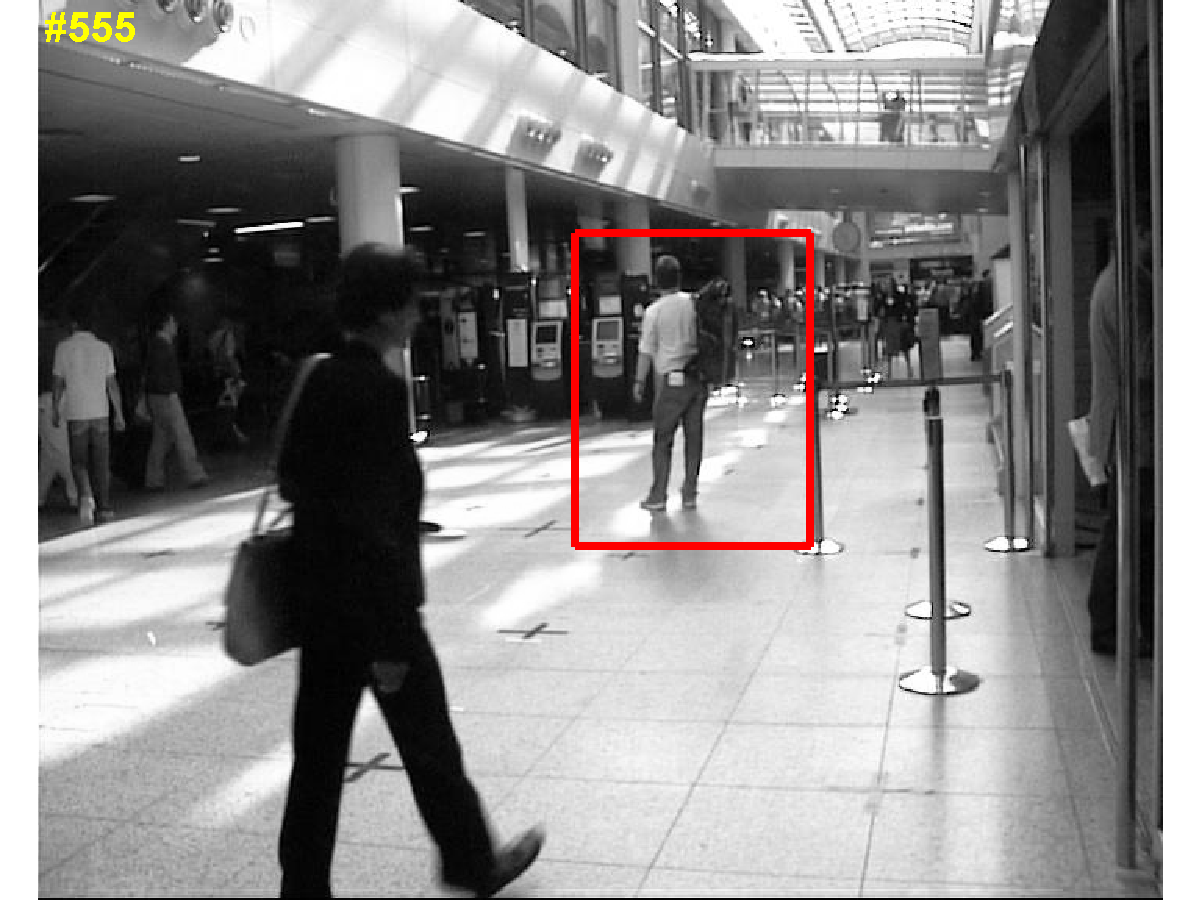} &
      \includegraphics[width=2in]{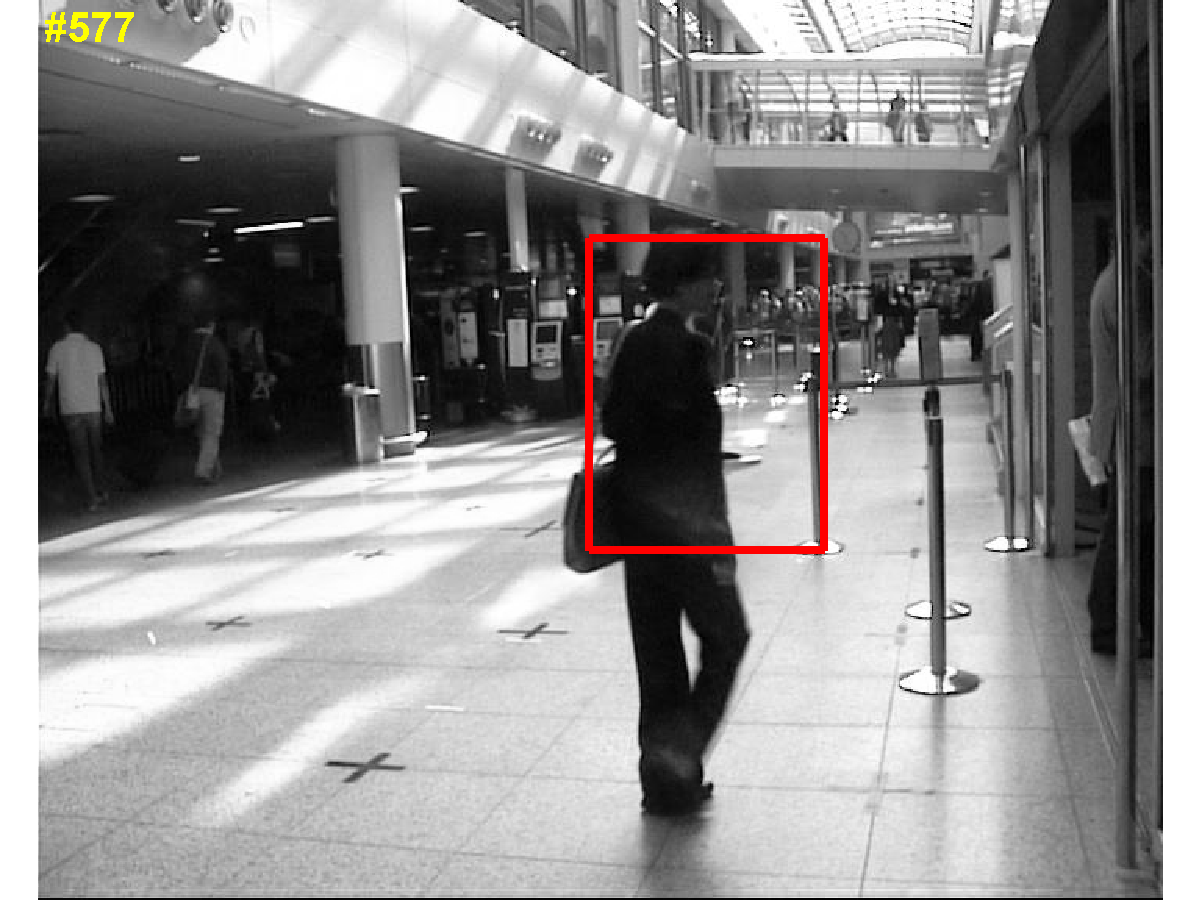} \\
       (a) &  (b) & (c) \\
      \hline
      \includegraphics[width=2in]{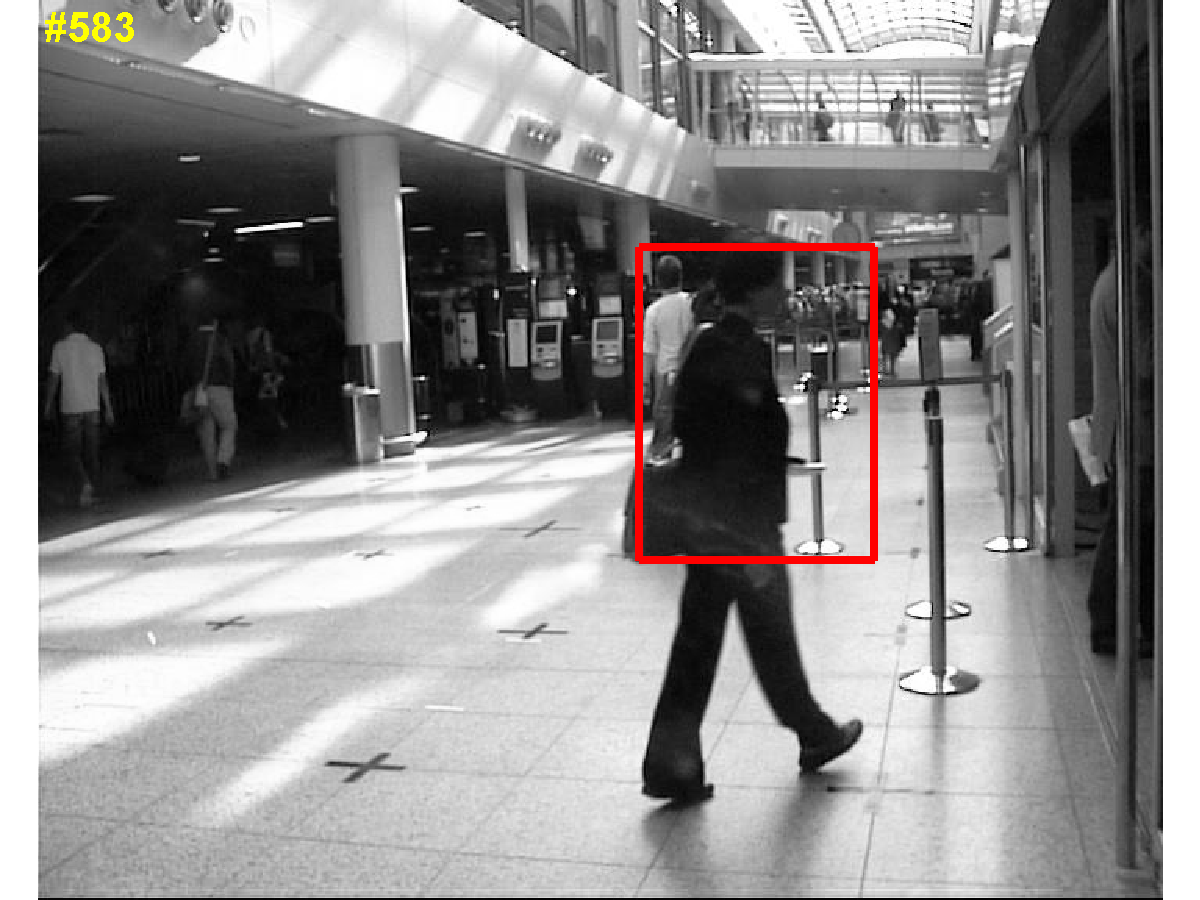} &
      \includegraphics[width=2in]{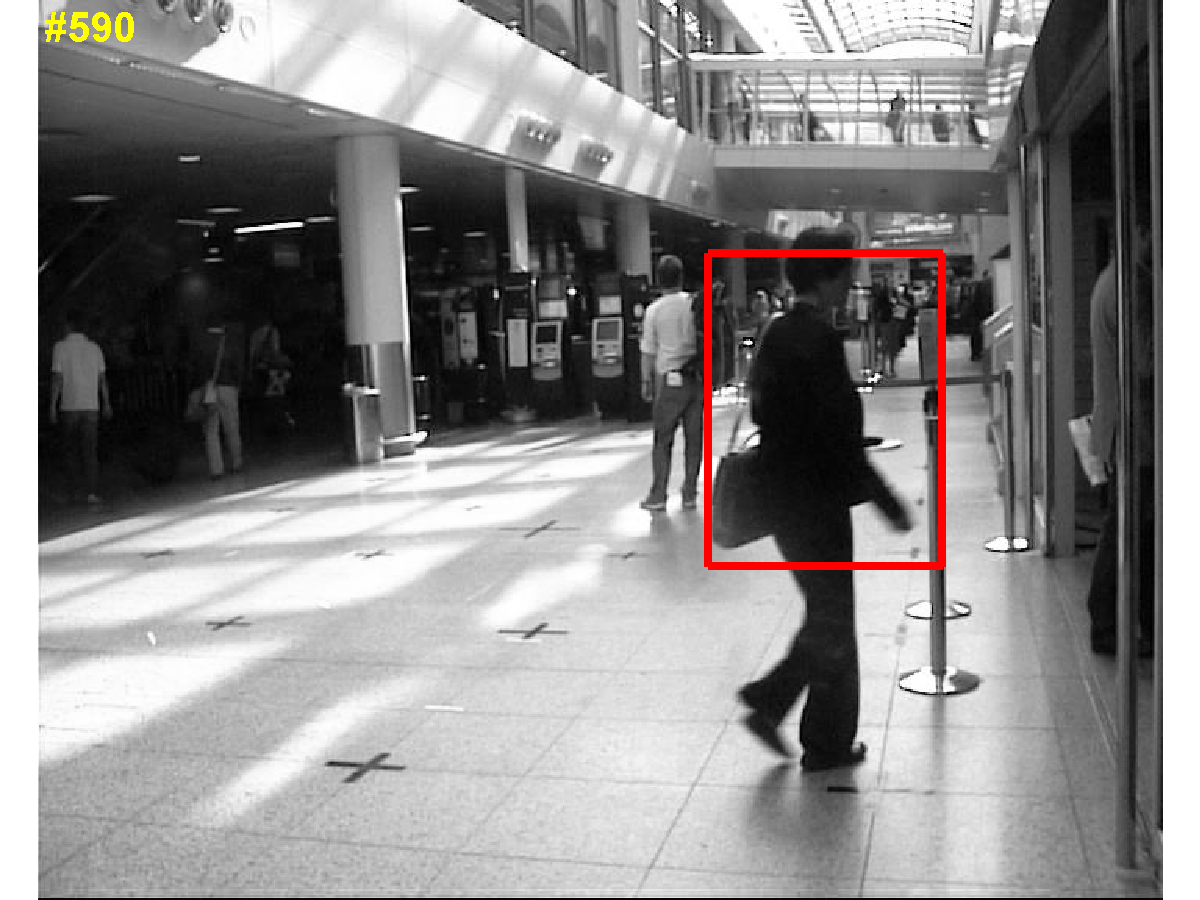} &
      \includegraphics[width=2in]{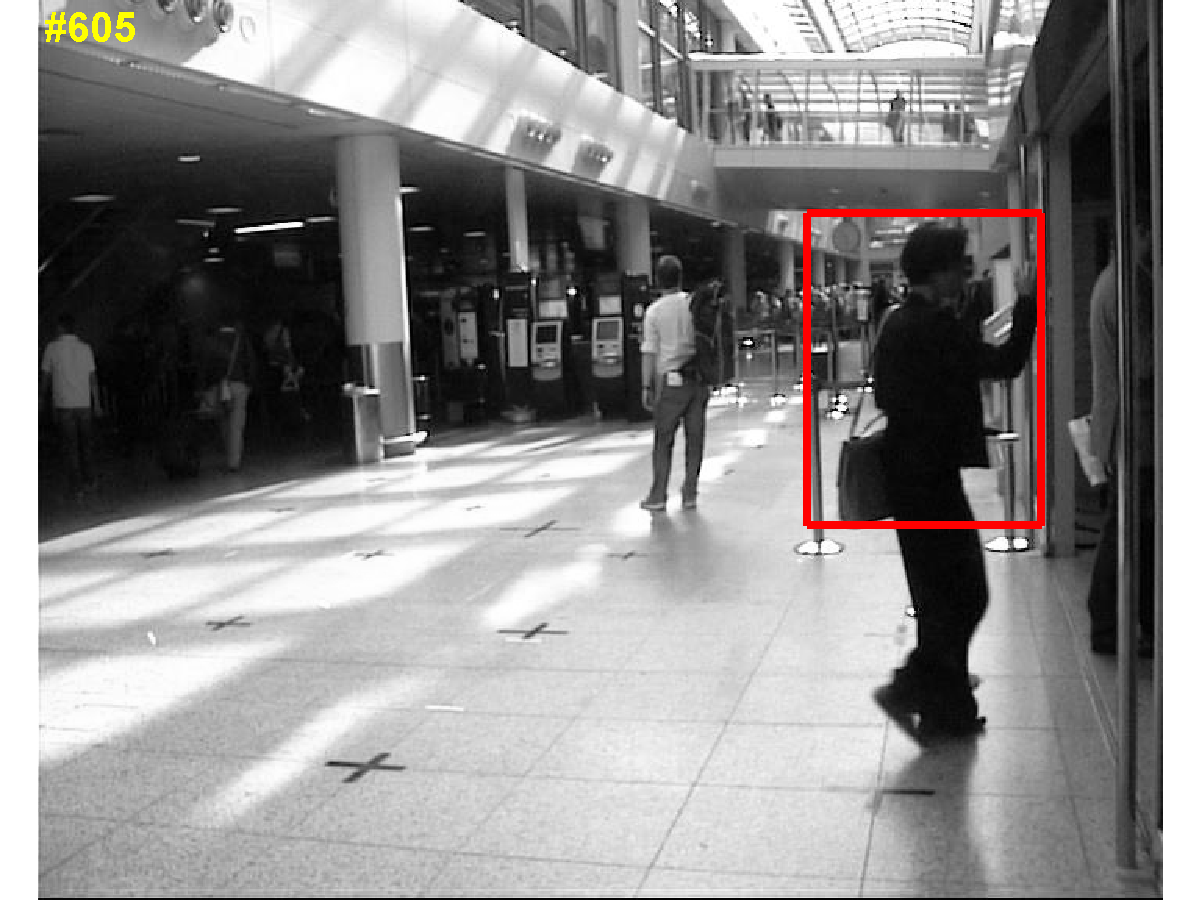}   \\
      \hline
      (d) &  (e) & (f) \\
      \hline
\end{tabular}
  \caption{Drifting problem of the real-time compressive tracking algorithm with a fixed learning parameter $\lambda=0.85$. When the being-tracked person is completely occluded by the passenger at frame 577 (sub-figure (c)), the discriminative model is updated according to the appearance of the passenger. Thus, a tracked loss occurs.  (the figure is best viewed in color)
  }
   \label{exp1}
\end{figure*}

\begin{figure*}[h!]
\begin{tabular}{|c|c|c|}
      \hline
      \includegraphics[width=2in]{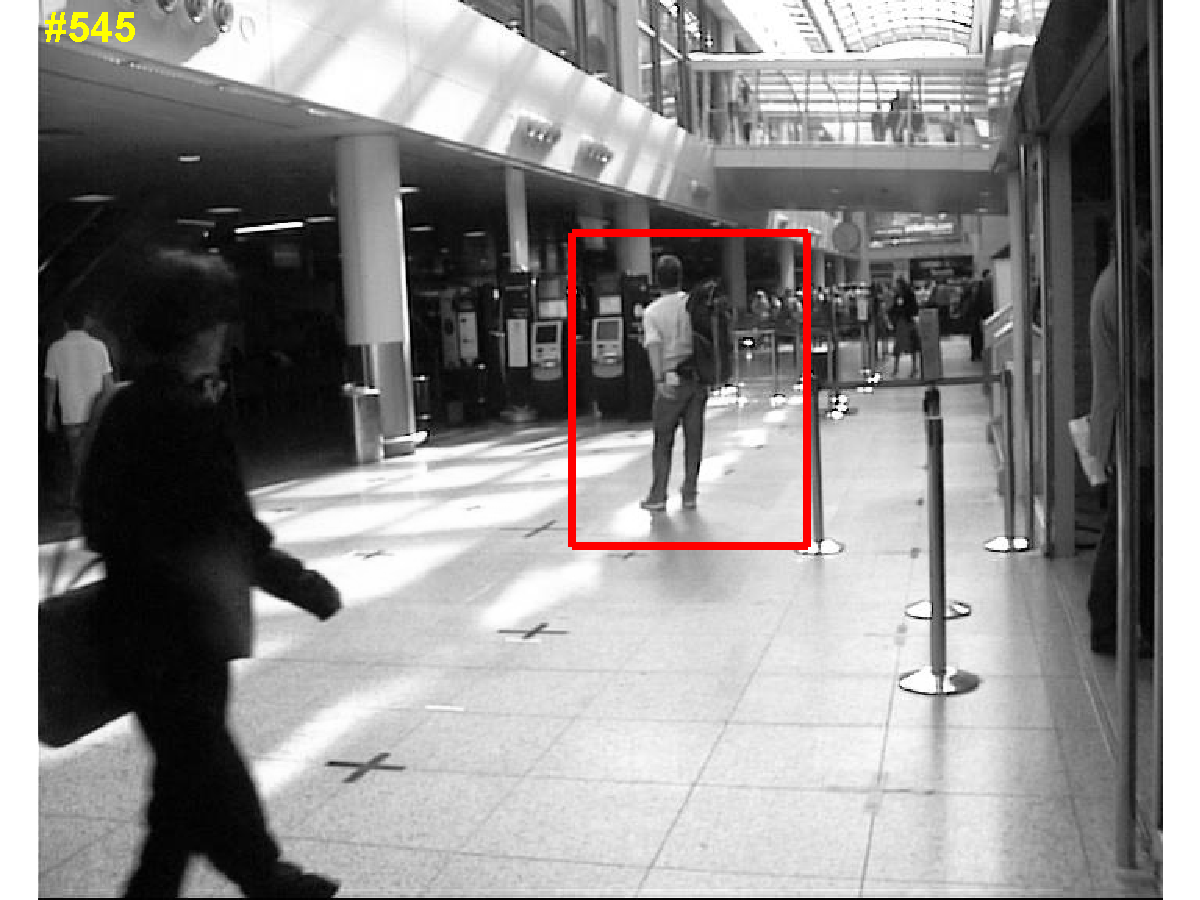} &
      \includegraphics[width=2in]{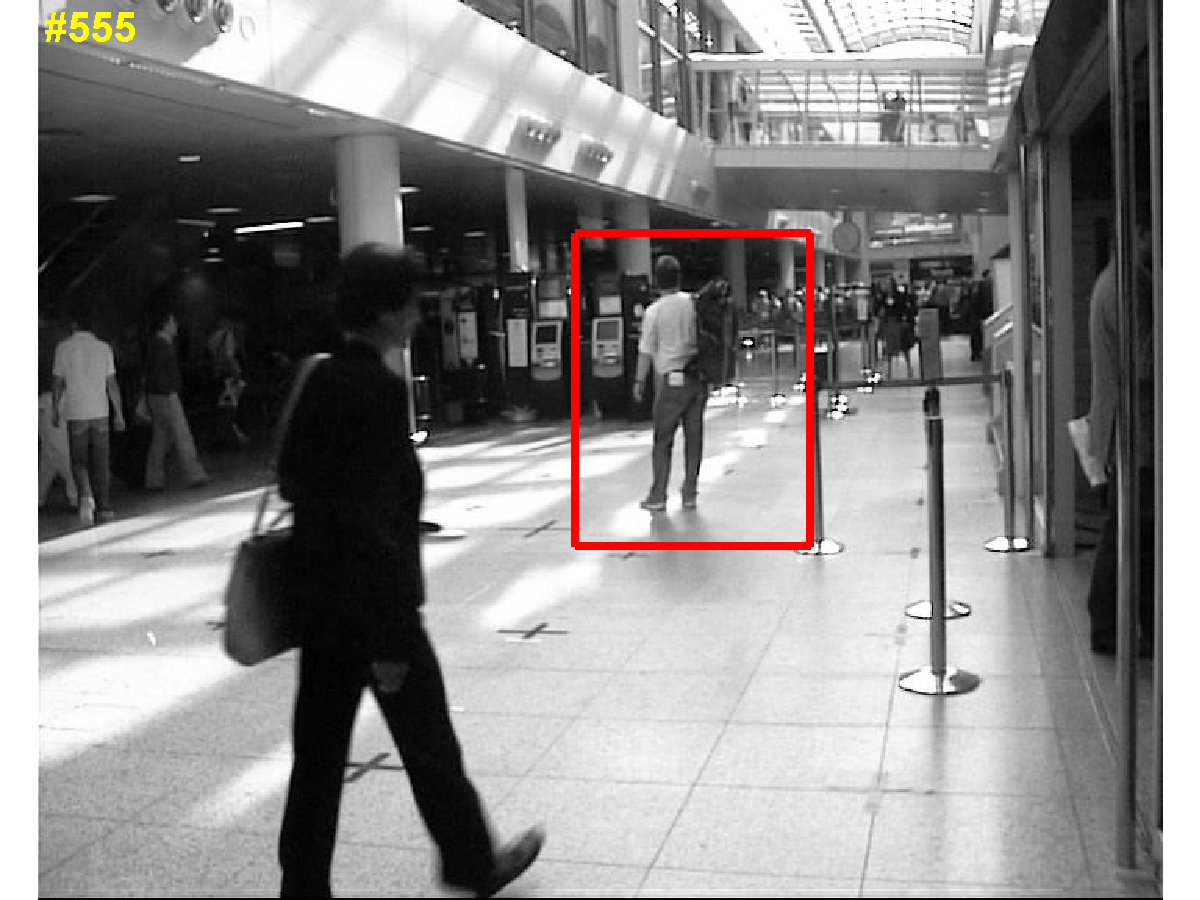} &
      \includegraphics[width=2in]{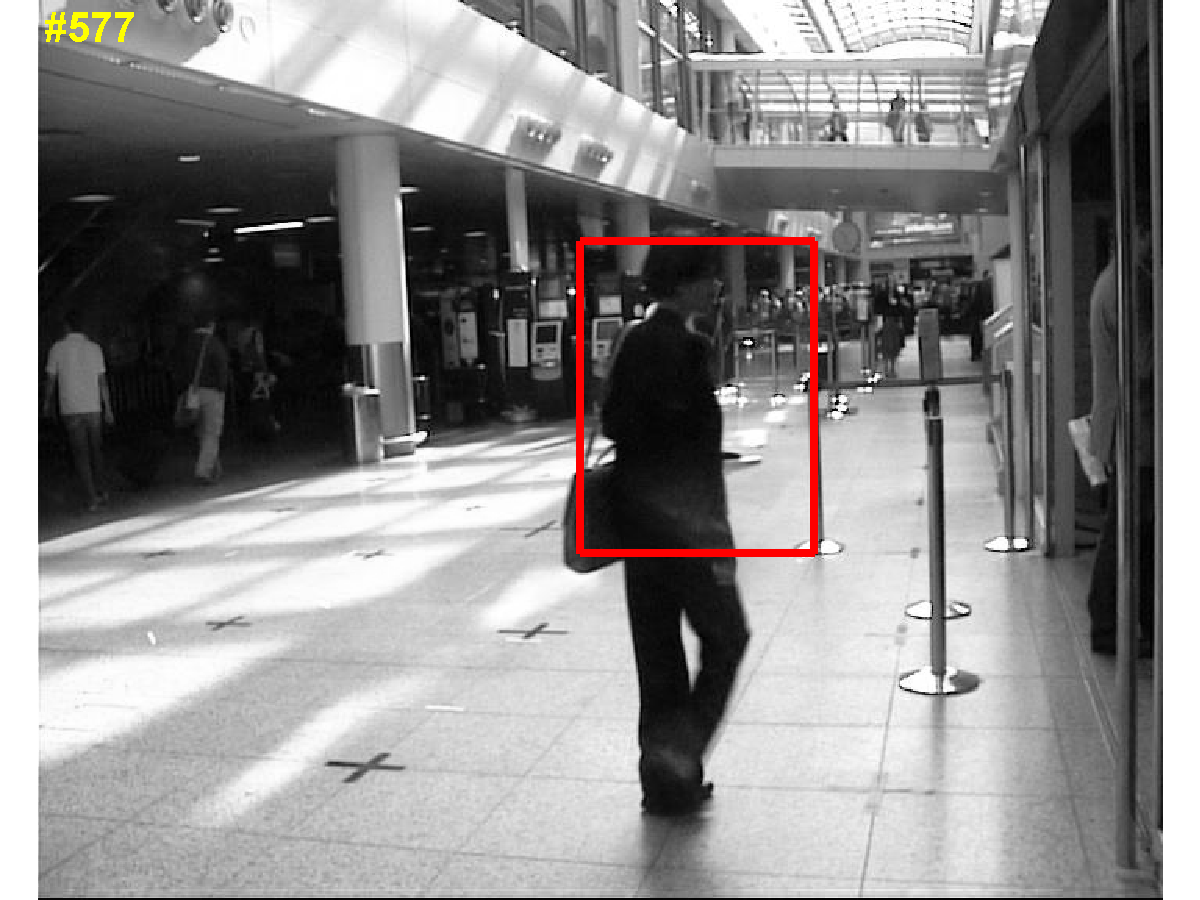} \\
      (a) &  (b) & (c) \\
      \hline
      \includegraphics[width=2in]{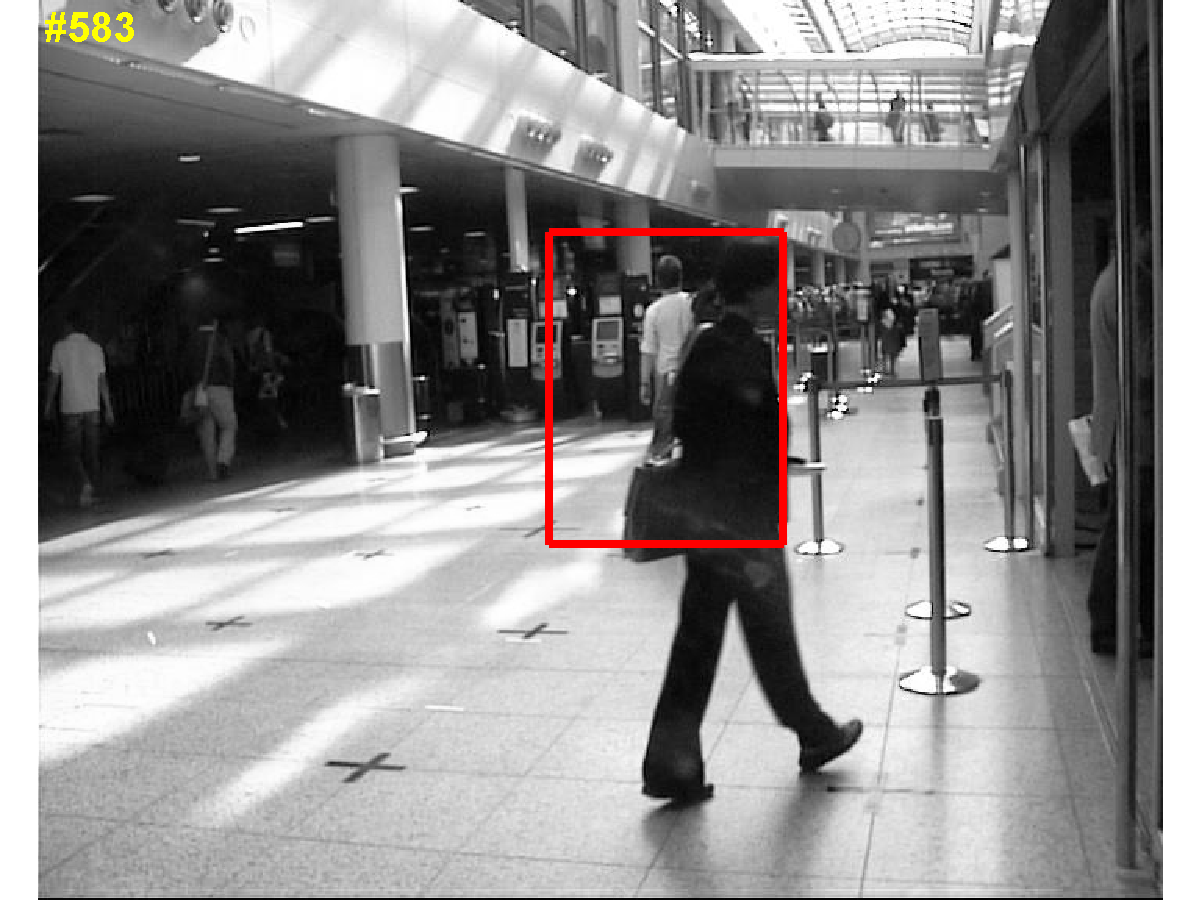} &
      \includegraphics[width=2in]{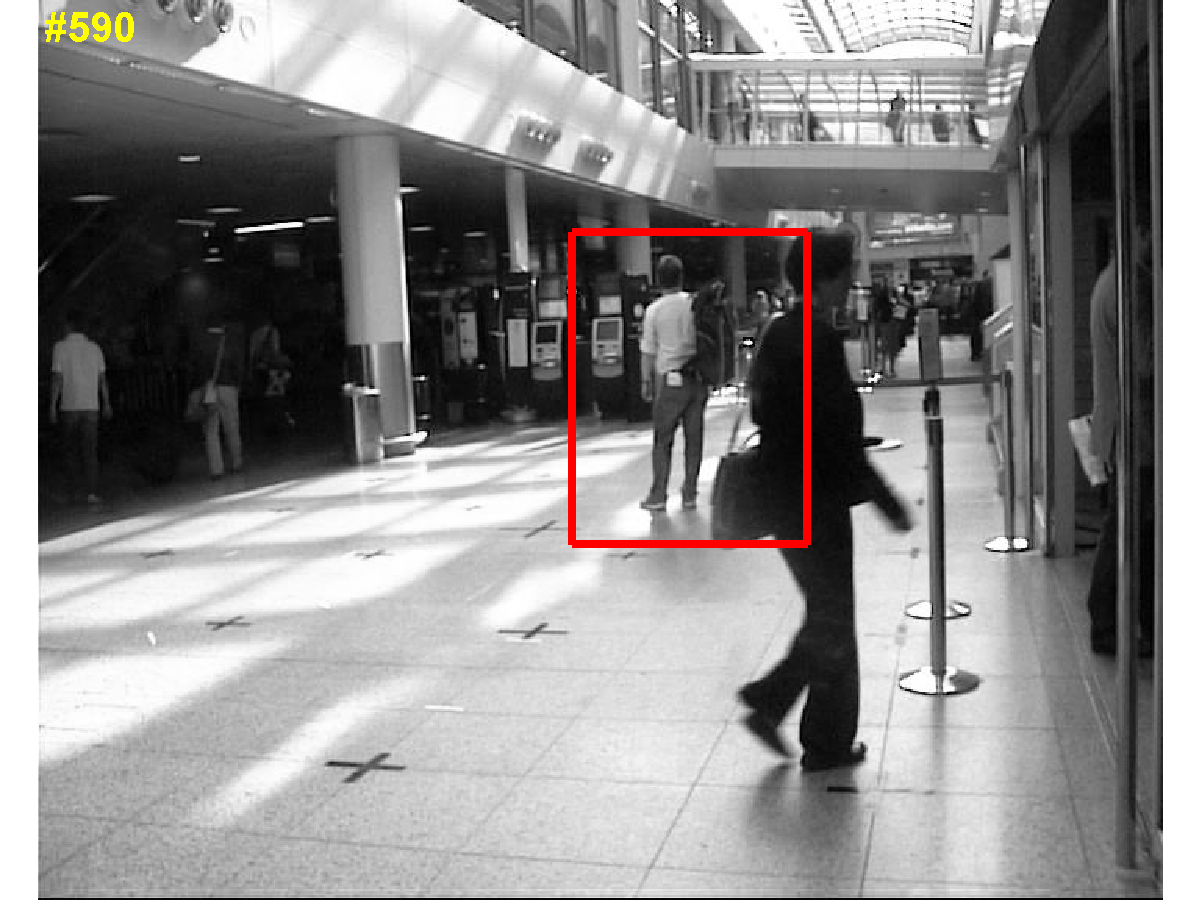} &
      \includegraphics[width=2in]{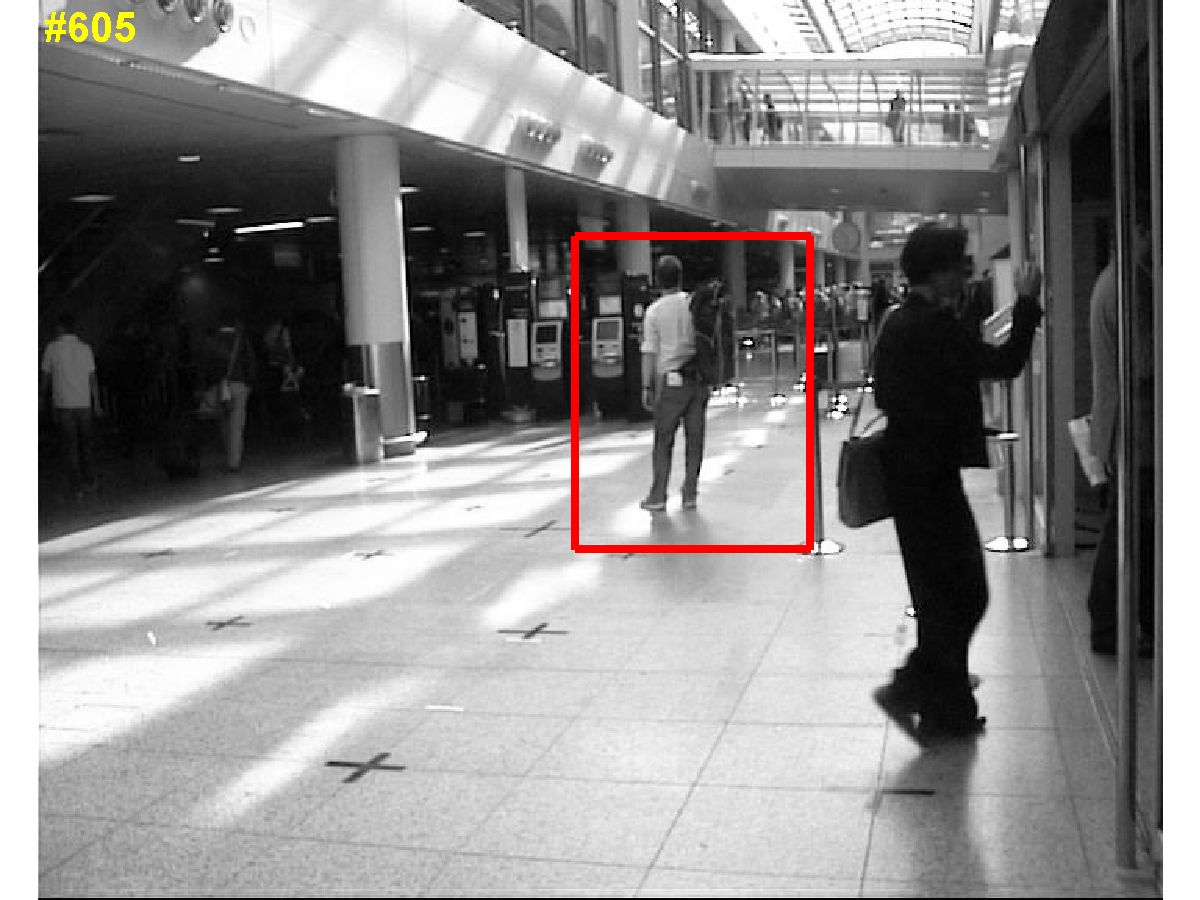}   \\
      \hline
      (d) &  (e) & (f) \\
      \hline
\end{tabular}
  \caption{Drifting is avoided by the proposed algorithm}
     \label{exp2}
\end{figure*}

We observe that the real-time compressive tracking systems with a fixed learning parameter in the prior art \cite{zzy12} do have the drifting problem as shown in Fig. \ref{exp1}. We show in Fig. \ref{exp2} that such drifting problem can be avoided by the algorithm proposed in this paper. The occlusion event around frame 577 is detected by our algorithm very clearly with the corresponding probability close to $1$.

\begin{figure*}[h!]
\begin{tabular}{|c|c|c|}
      \hline
      \includegraphics[width=2in]{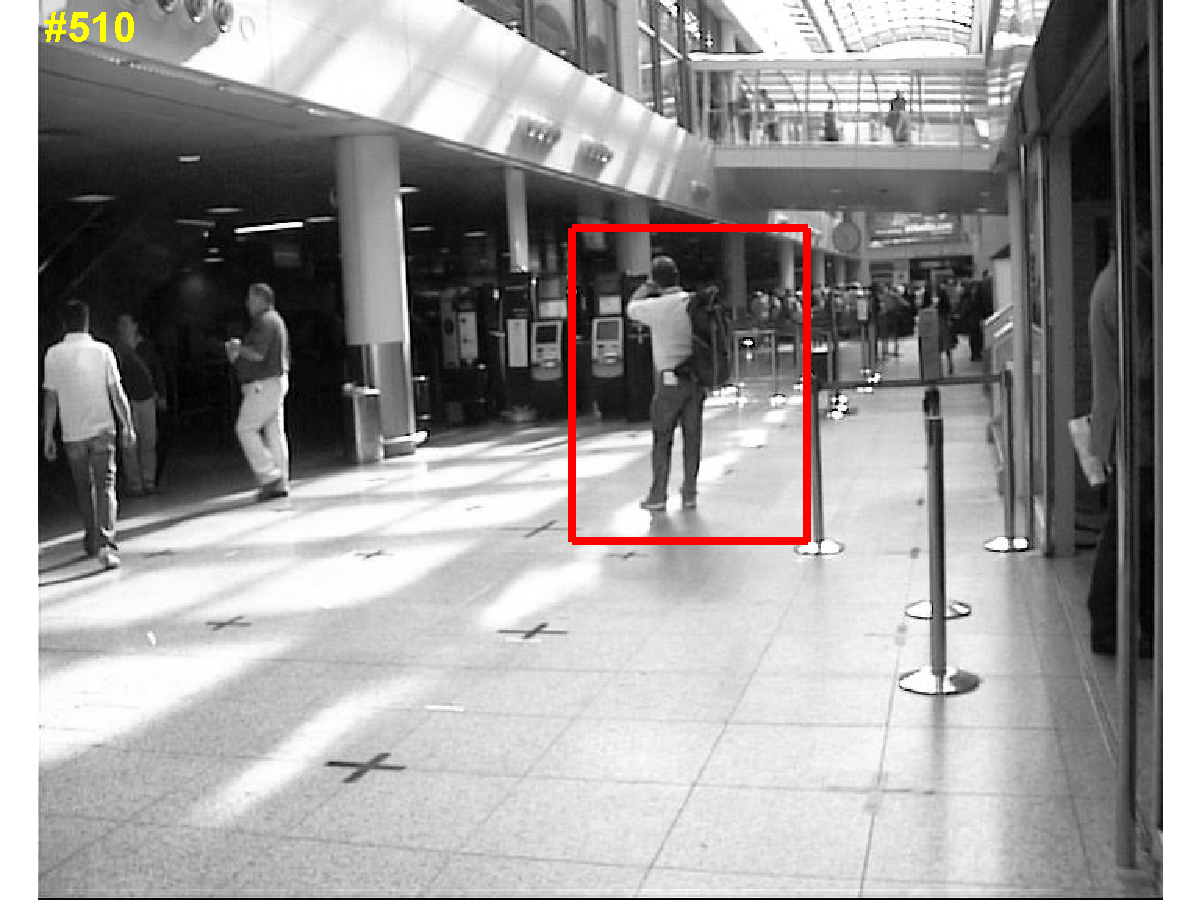} &
      \includegraphics[width=2in]{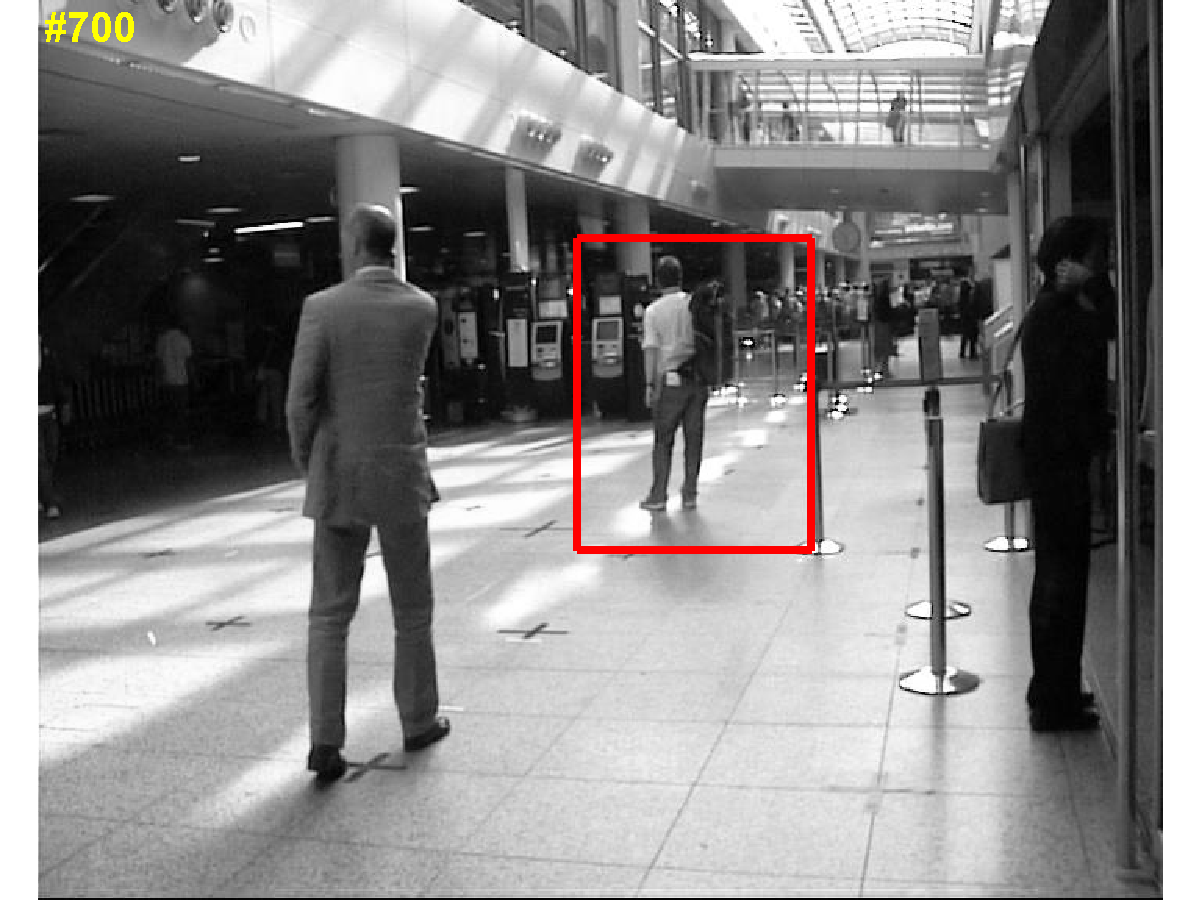} &
      \includegraphics[width=2in]{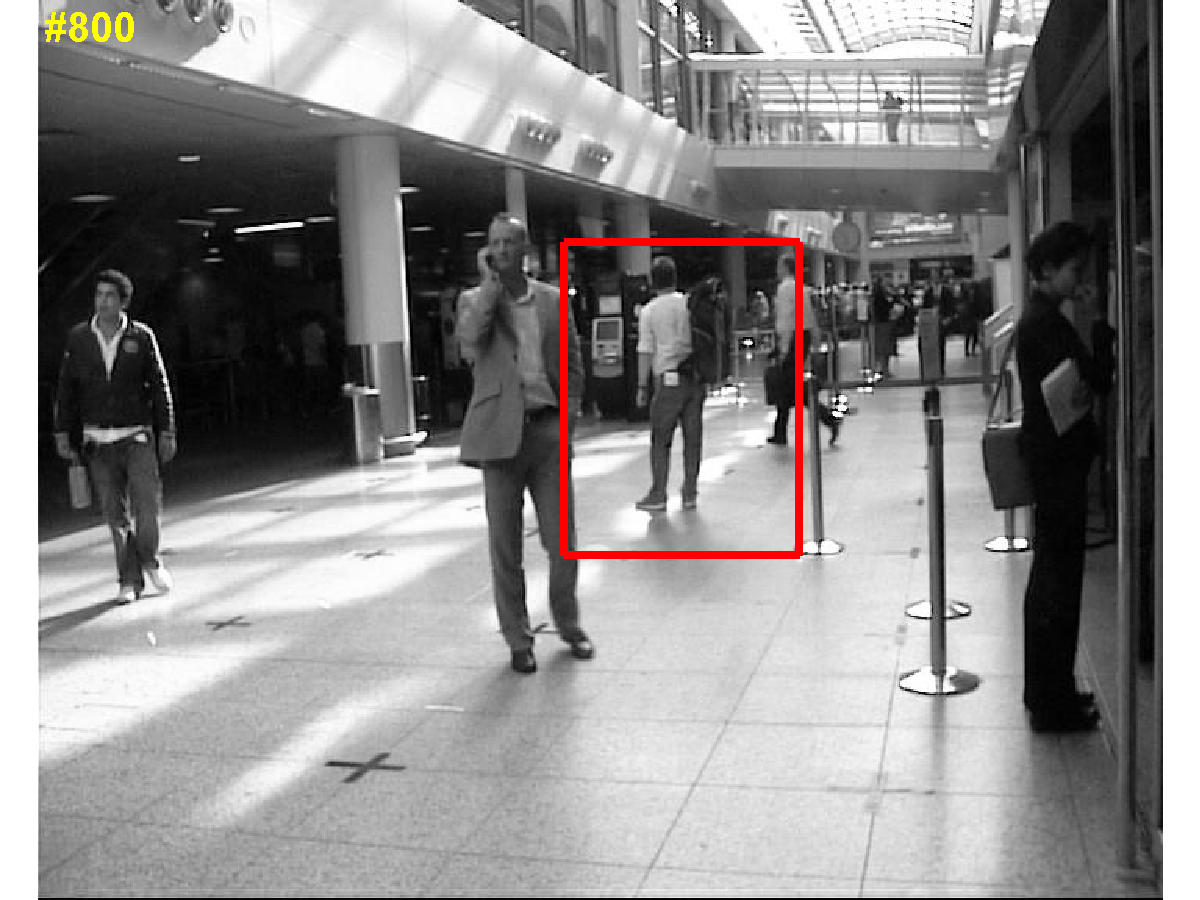} \\
      (a) &  (b) & (c) \\
      \hline
      \includegraphics[width=2in]{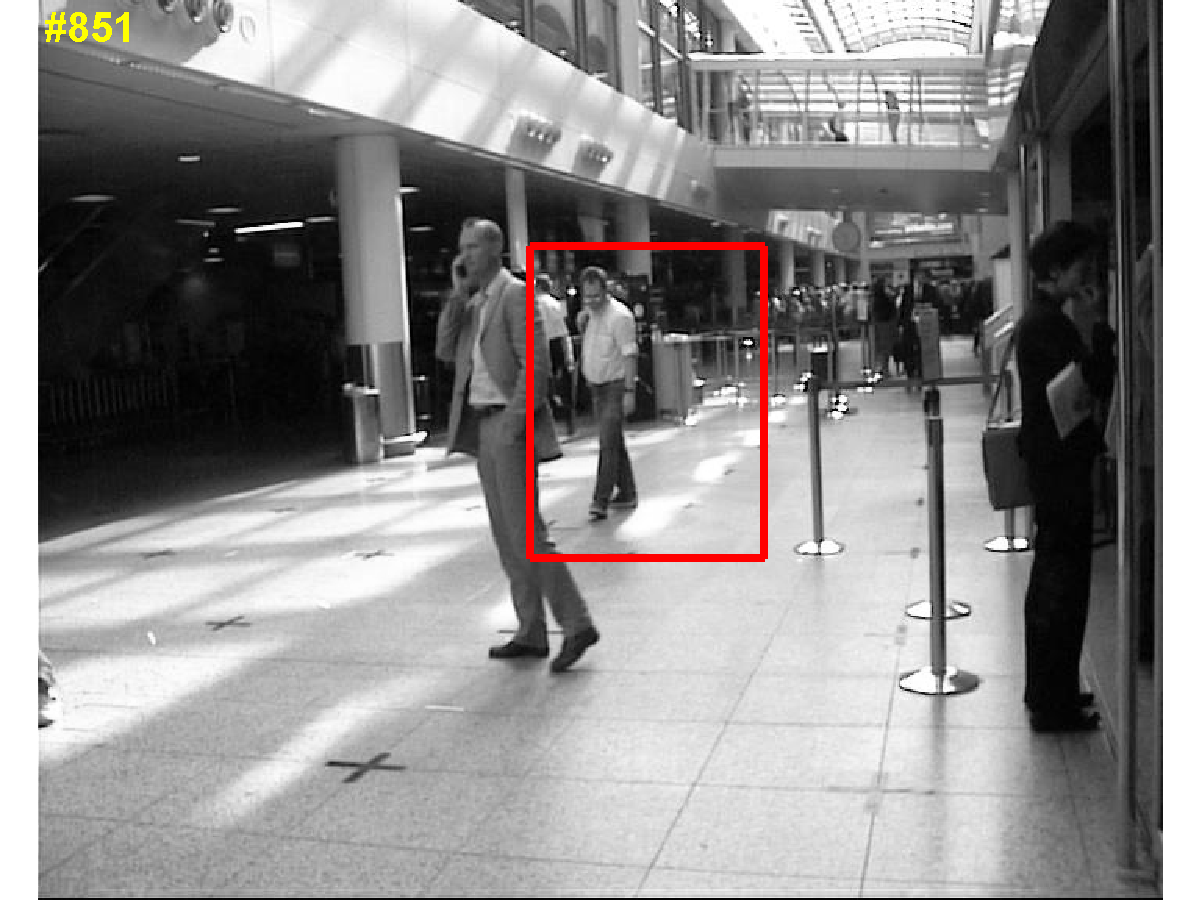} &
      \includegraphics[width=2in]{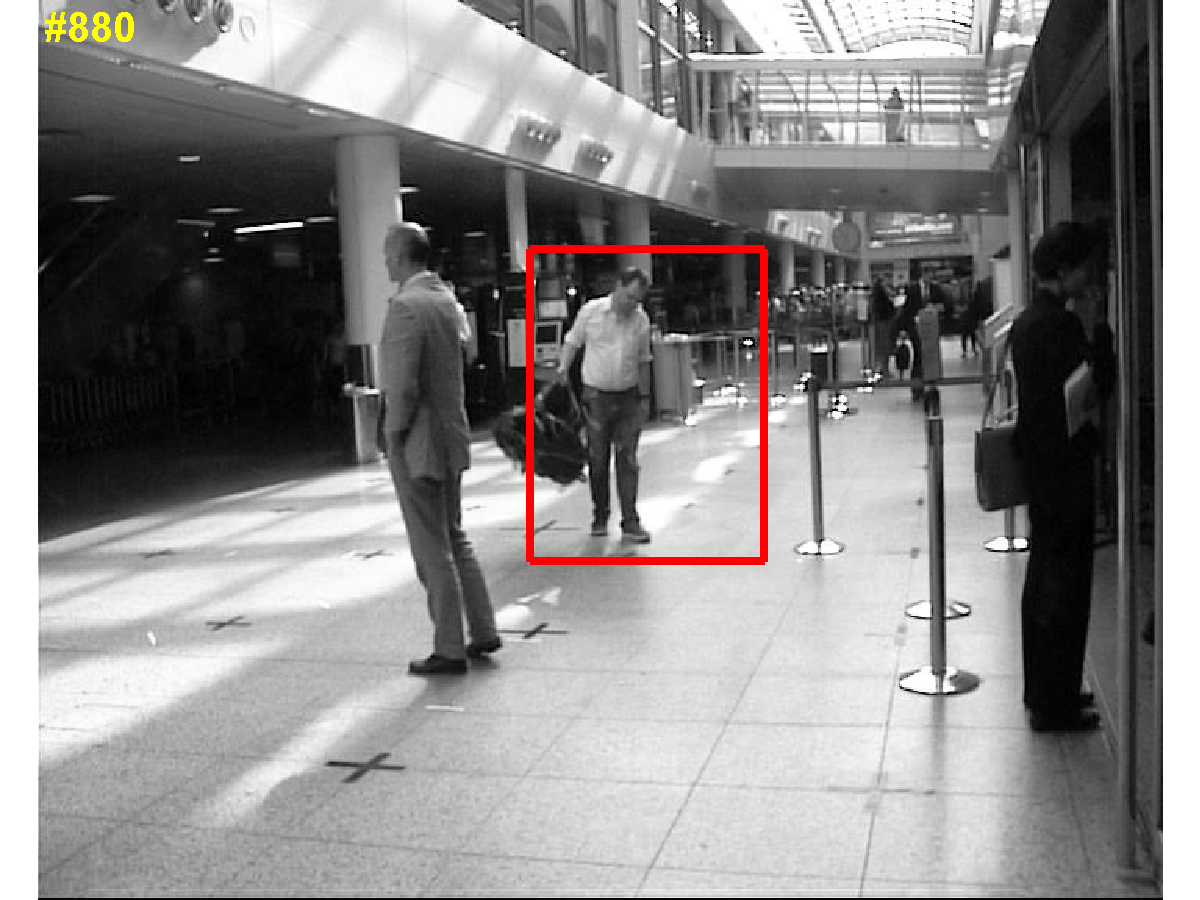} &
      \includegraphics[width=2in]{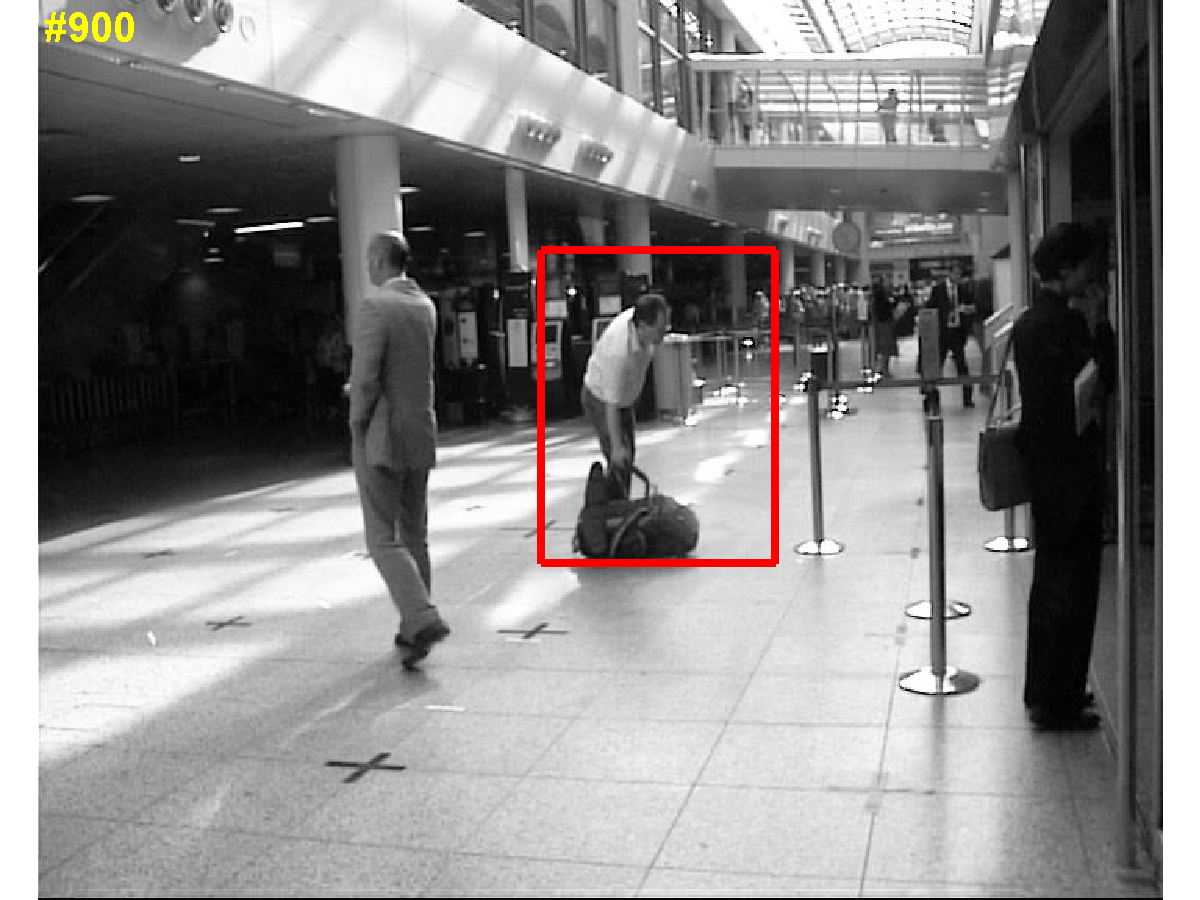}   \\
      \hline
      (d) &  (e) & (f) \\
      \hline
\end{tabular}
  \caption{Detecting sudden-appearance change }
     \label{exp3}
\end{figure*}

\begin{figure*}[h!]
\begin{tabular}{|c|c|c|}
      \hline
      \includegraphics[width=2in]{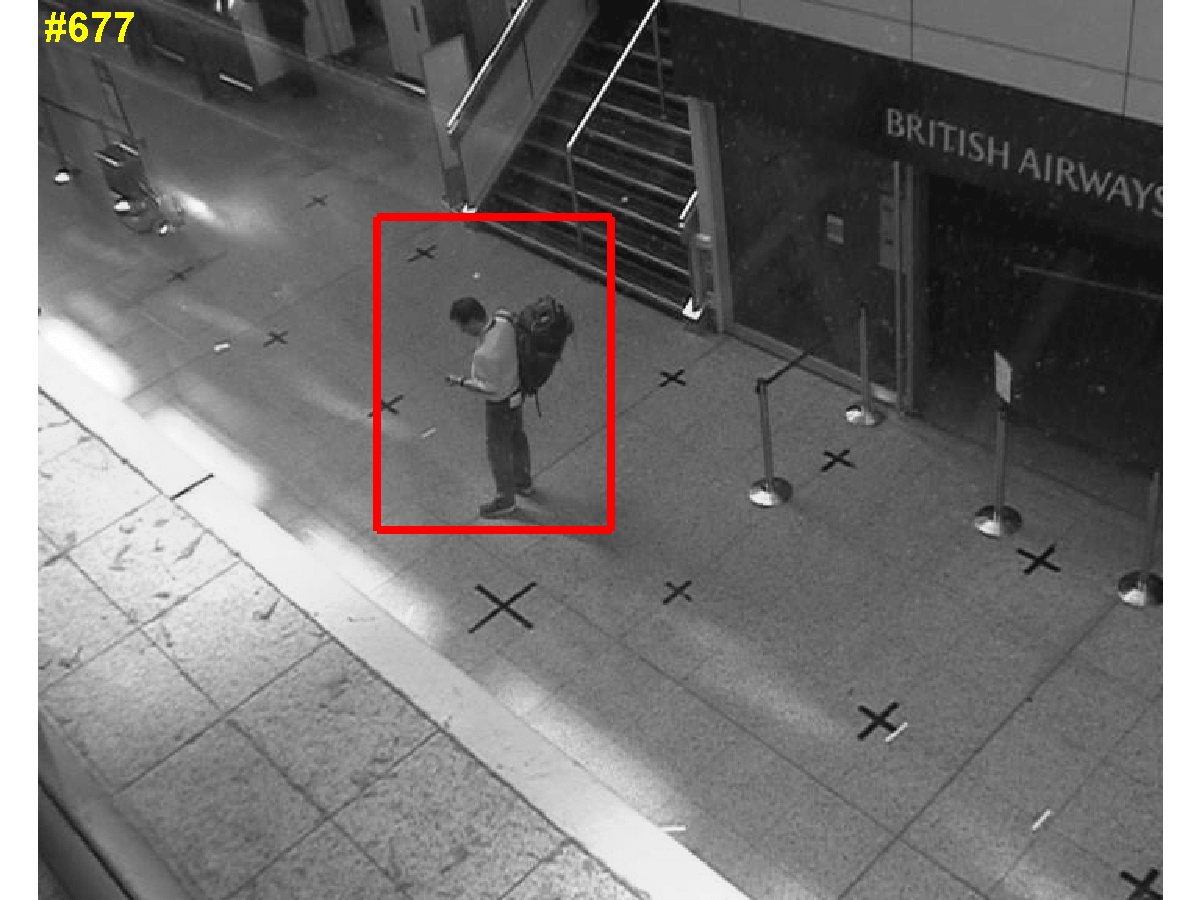} &
      \includegraphics[width=2in]{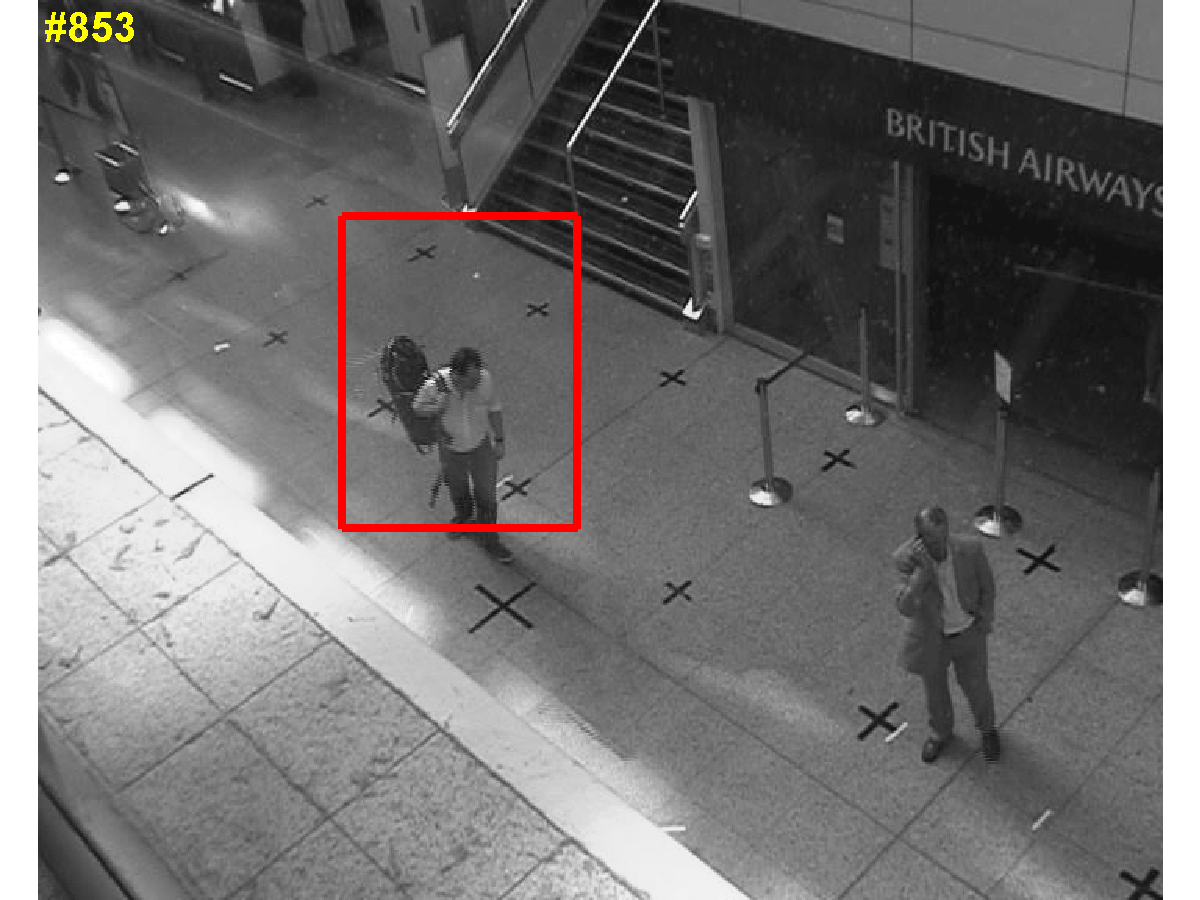} &
      \includegraphics[width=2in]{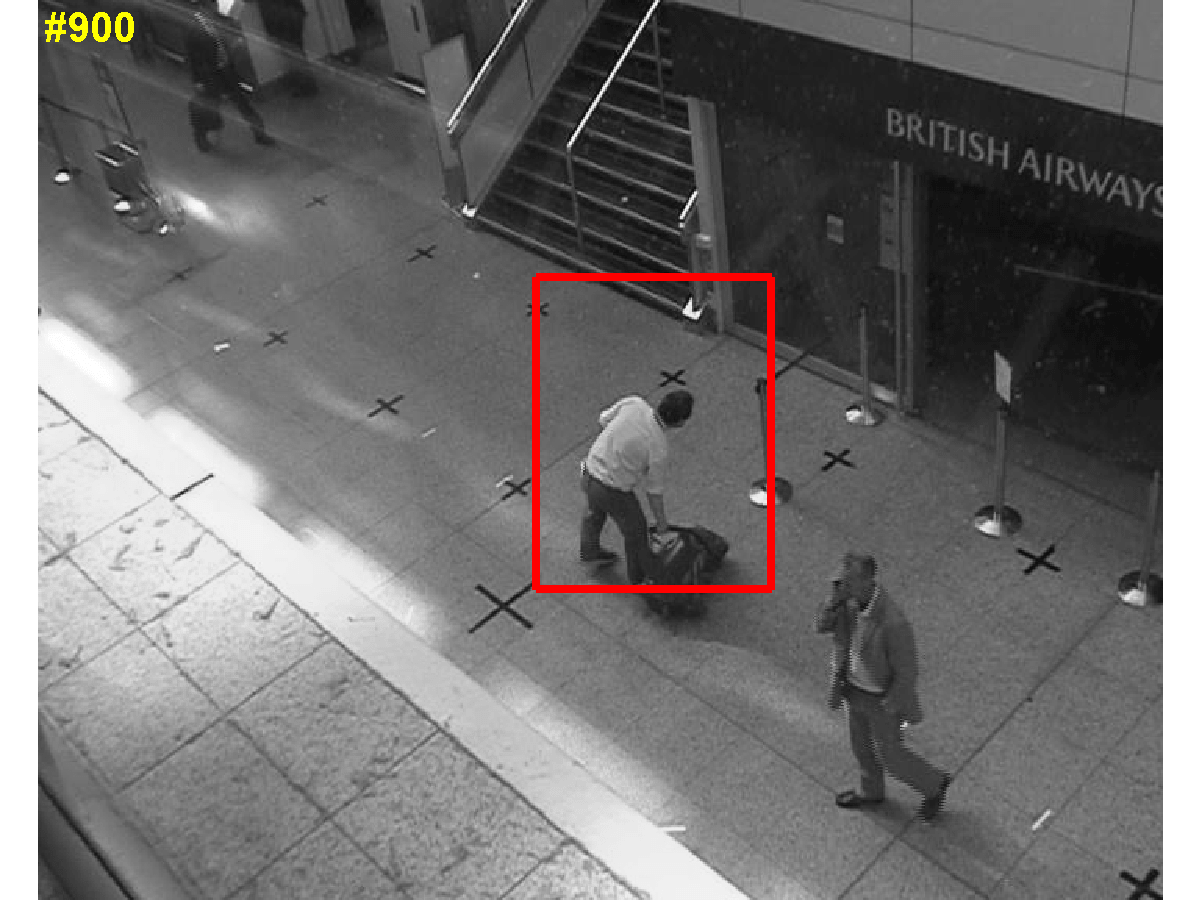} \\
      (a) &  (b) & (c) \\
      \hline
      \includegraphics[width=2in]{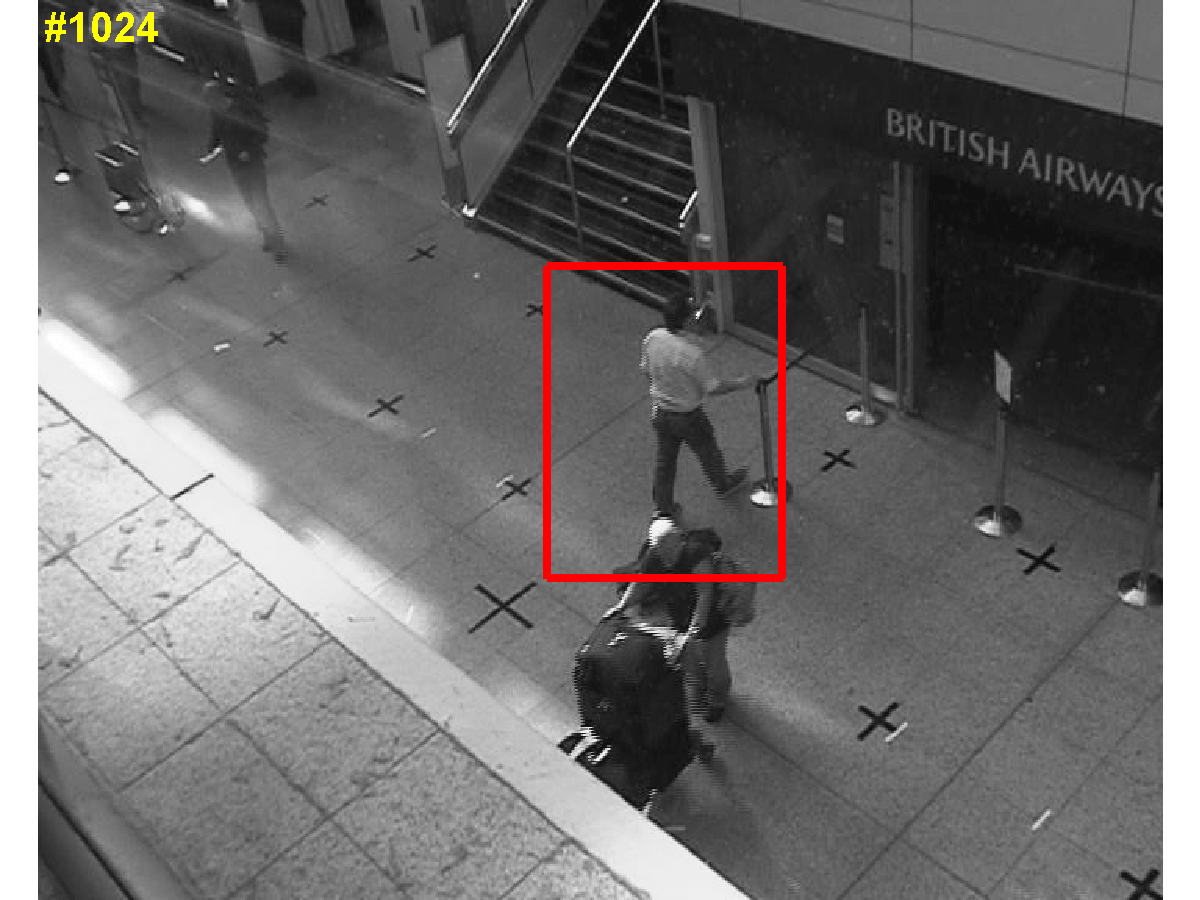} &
      \includegraphics[width=2in]{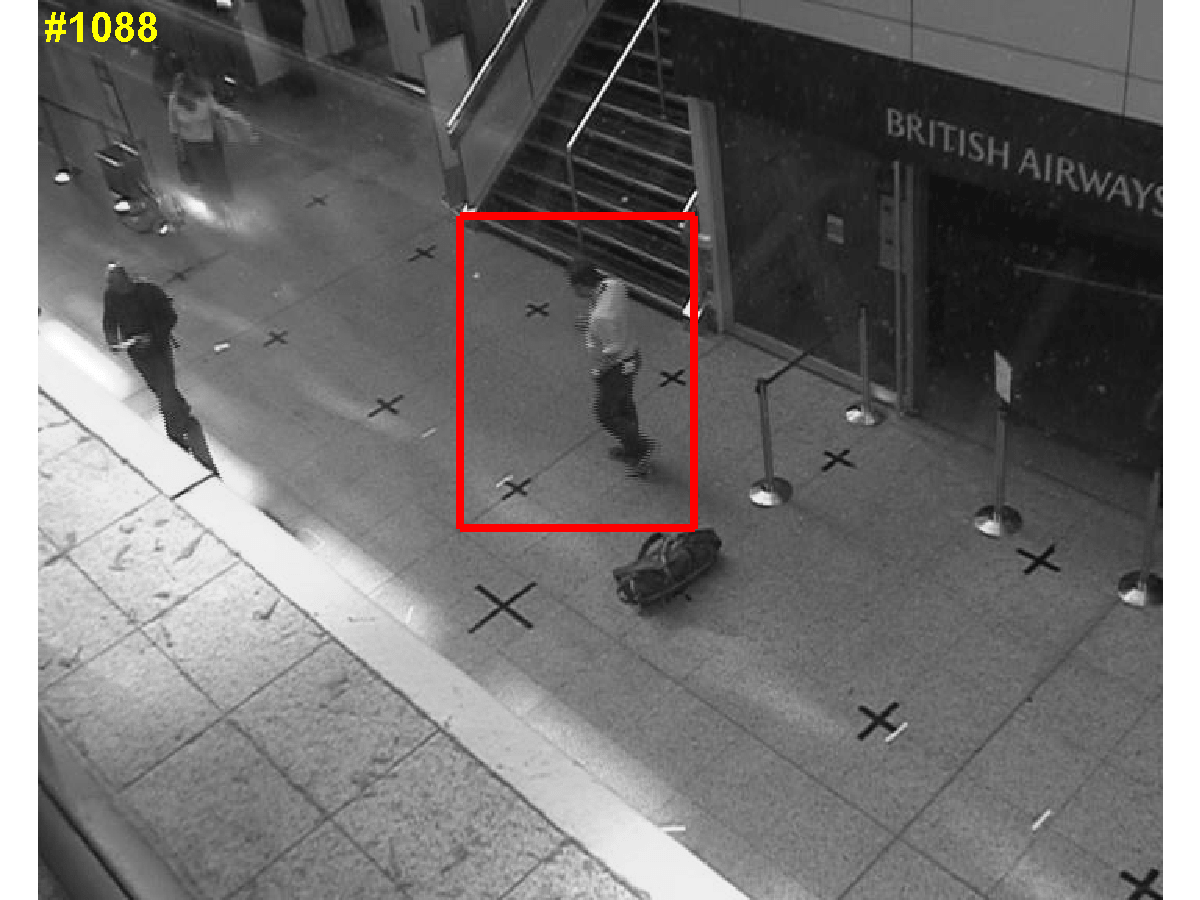} &
      \includegraphics[width=2in]{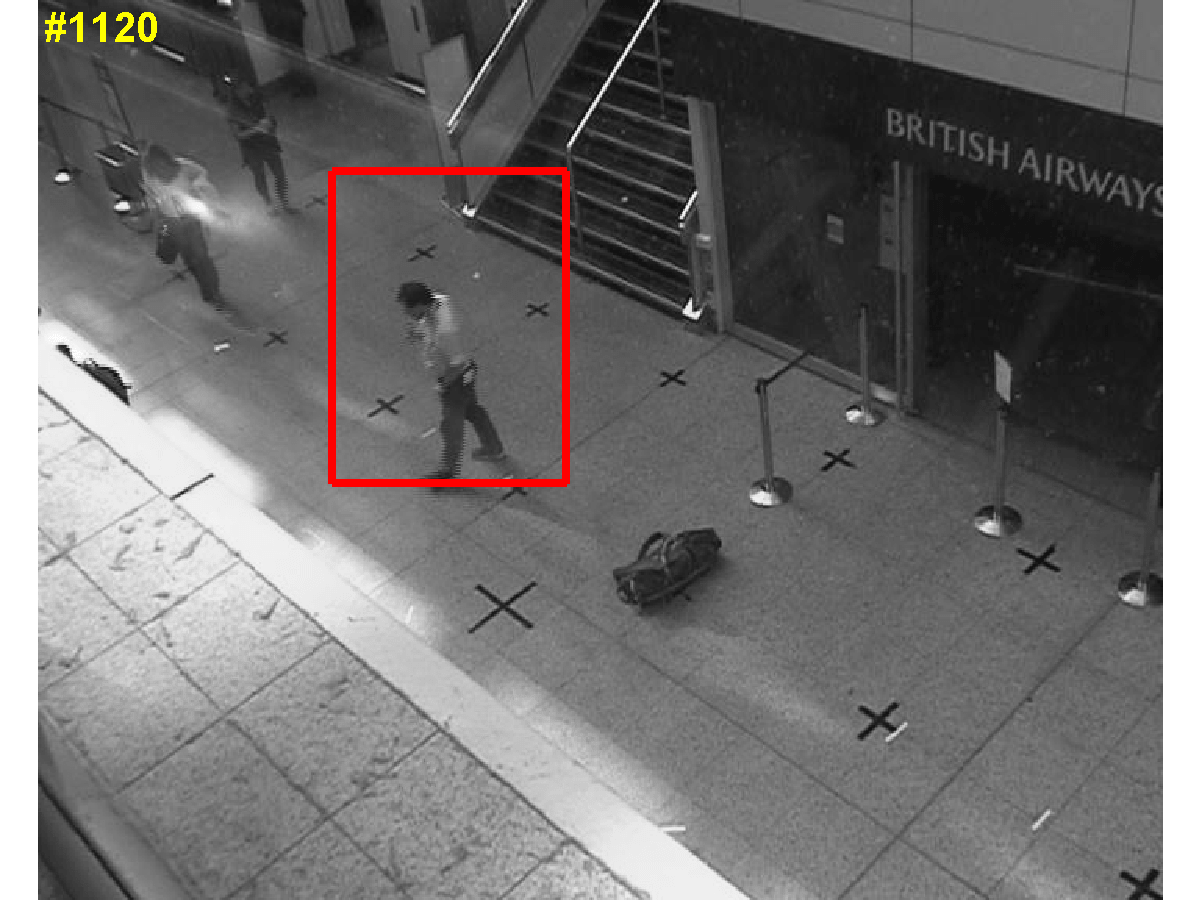}   \\
      \hline
      (d) &  (e) & (f) \\
      \hline
\end{tabular}
  \caption{Detecting sudden-appearance change at another camera}
     \label{exp4}
\end{figure*}

The proposed algorithm also detects the unusual behavior of the being-tracked person at frame 851 (with the corresponding probability higher than $0.9$). The proposed algorithm is able to track the where-about of the suspicious person as shown in Figs. \ref{exp3}, and \ref{exp4}, where each figure shows the tracking results at one camera. 

In all the above experimental results, the tracking results are shown as red bounding boxes. And the frame indexes are labelled in all the images.

\section{Conclusion}

\label{sec_conclusion}

The paper discusses a visual tracking algorithm to detect sudden-appearance-change and occlusions. By experimental results, we show that the proposed algorithm can reliably detect the sudden-appearance-change and occlusion events. Such reliable estimations can also be used to avoid the drifting problems.

\section*{Acknowledgment}

This research was originally submitted to \mbox{Xinova, LLC} by the author in response to a Request for Invention. It is among several submissions that Xinova has chosen to make available to the wider community. The author wishes to thank Xinova, LLC for their funding support of this research. More information about Xinova, LLC is available at www.xinova.com.

\bibliographystyle{IEEEtran}
\bibliography{the_bib}

\end{document}